\documentclass[sigconf]{acmart}

\AtBeginDocument{%
  }

\usepackage{tcolorbox}
\usepackage{graphicx}
\usepackage{multirow}
\usepackage{alltt}
\usepackage{booktabs} 
\begin{document}

\title{IMPROVING MULTIMODAL LLM'S ABILITY IN GEOMETRY PROBLEM SOLVING, REASONING, AND MULTISTEP SCORING}

%
\author{Avinash Anand}
\email{avinasha@iiitd.ac.in}
\affiliation{%
  \institution{IIIT Delhi}
  \country{India}
}
\author{Raj Jaiswal}
\email{Jaiswalp@iiitd.ac.in}
\affiliation{%
  \institution{IIIT Delhi}
  \country{India}
}
\author{Abhishek  Dharmadhikari}
\email{abhishekdharmadhikari25@gmail.com}
\affiliation{%
  \institution{MIDAS LAB}
  \country{India}
}
\author{Atharva	Marathe}
\email{atharvamarathe8@gmail.com}
\affiliation{%
  \institution{MIDAS LAB}
  \country{India}
}
\author{Harsh Parimal Popat}
\email{harsh21048@iiitd.ac.in}
\affiliation{%
  \institution{IIIT Delhi}
  \country{India}
}
\author{Harshil Mital}
\email{harshil21050@iiitd.ac.in}
\affiliation{%
  \institution{IIIT Delhi}
  \country{India}
}
\author{Kritarth Prasad}
\email{kritarth20384@iiitd.ac.in}
\affiliation{%
  \institution{IIIT Delhi}
  \country{India}
}

\author{Rajiv Ratn Shah}
\email{rajivratn@iiitd.ac.in}
\affiliation{%
  \institution{IIIT Delhi}
  \country{India}
}
\author{Roger Zimmermann}
\email{rogerz@comp.nus.edu.sg}
\affiliation{%
  \institution{NUS}
  \country{Singapore}
}







\renewcommand{\shortauthors}{Anand et al.}

\begin{abstract}
This paper presents GPSM4K, a comprehensive geometry multimodal dataset tailored to augment the problem-solving capabilities of Large Vision Language Models (LVLMs).
GPSM4K  encompasses 2157 multimodal question-answer pairs manually extracted from mathematics textbooks spanning grades 7-12 and is further augmented to 5340 problems, consisting of both numerical and theorem-proving questions.
In contrast to PGPS9k, Geometry3K, and Geo170K which feature only objective-type questions, GPSM4K offers detailed step-by-step solutions in a consistent format, facilitating a comprehensive evaluation of problem-solving approaches. This dataset serves as an excellent benchmark for assessing the geometric reasoning capabilities of LVLMs. Evaluation of our test set shows that there is scope for improvement needed in open-source language models in geometry problem-solving. Finetuning on our training set increases the geometry problem-solving capabilities of models. Further, We also evaluate the effectiveness of techniques such as image captioning and Retrieval Augmentation generation (RAG) on model performance. We leveraged LLM to automate the task of final answer evaluation by providing ground truth and predicted solutions. This research will help to assess and improve the geometric reasoning capabilities of LVLMs.
\end{abstract}





\maketitle

\section{Introduction}
Geometry problems, characterized by their intricate relationship between textual descriptions and visual representations, are exemplary benchmarks for evaluating multimodal numerical reasoning capabilities. Despite the longstanding recognition of automatic geometric problem-solving as a critical benchmark in AI research, the availability of suitable datasets is limited. This dearth stems from the inherent complexity and diverse array of information that geometry problems involve.

Early on, there were small-scale datasets that relied heavily on manual annotation to assist geometric problem-solving models \cite{seo2014diagram} and GEOS \cite{seo2015solving} is a dataset of SAT  plane geometry questions with 186 questions. Then the Geometry3k \cite{lu2021inter} was proposed, which not only increased the quantity of data but also enriched the geometric problem types in terms of geometric shapes and variable operators. At the same time, a larger and more diverse dataset, GeoQA \cite{chen2021geoqa}, included clear annotations of the problem-solving process. It improved the universality and interpretability of multi-modal numerical reasoning. Language ability does not equal 'thinking' or 'reasoning' in LLMs. One of the long-term goals of artificial intelligence is to develop machines with the ability to reason mathematically.

Existing geometry datasets like Geometry3K, PGPS9K lack a diverse range of problem statements, including Numerical Answer Questions and Theorem Proving Questions, essential for secondary-level education. To address this issue, one of the most direct and effective approaches is to enhance current Multimodal Large Language Models(MLLMs) by augmenting them with data containing high-quality descriptions of geometric information. However, the primary obstacle remains the small scale of existing geometric problem-solving datasets, typically offering only a few question-answer pairs. These datasets' lack of geometric image descriptions and a narrow scope of problem-solving techniques limit models' understanding of basic geometric concepts and their problem-solving abilities. 

To address the limitations of existing datasets, we develop a comprehensive multimodal geometry dataset named {GPSM4K }. GPSM4K consists of approximately 4,000 geometric image-caption pairs and question-answer pairs, extending the diversity and complexity of problems compared to Geometry3K.


This paper investigates several key aspects of our research study
\begin{itemize}
    \item \textbf{Improvement in LLM understanding:}  We examine how the integration of image captioning aids Large Language Models (LLMs) in comprehending images. The findings suggest that while image captioning can bridge some gaps between visual content and linguistic interpretation, its effectiveness depends on how advanced the captioning algorithms are and how relevant the captions are to the context as discussed in section. 
    \item \textbf{Visual Model capabilities :} The exploration into the Visual Encoder model's interaction with Large Language Models (LLMs) reveals that, while the Visual Encoder is pivotal for processing visual data, it may inadvertently constrain the LLM's capability if not adequately aligned with the LLM’s linguistic processing capabilities. This misalignment could limit the model's overall efficacy in interpreting and solving geometry problems.
    \item \textbf {Step By Step Solution Framework and Diverse Question Types}: Unlike traditional datasets, GPSM4K provides two solution versions for each problem: the textbook's original solution and a reconstructed step-by-step version via Gemini Pro Vision. This structured approach facilitates a more detailed and methodical problem-solving process, allowing for clearer understanding and easier debugging of the solution path. Additionally, GPSM4K includes not only multiple-choice questions but also Numerical Answer Questions and Theorem Proving Questions.
\end{itemize}

Our key contributions are 
\begin{itemize}
    \item The introduction of GPSM4K, a diverse and extensive benchmark for geometry problem-solving, which includes question-answer pairs
    \item A novel dataset structure featuring both original and detailed step-by-step solutions, alongside a variety of question types
    \item Evaluation of image caption integration in multimodal geometry figures within GPSM4K to enhance contextual understanding of visual data and impact of Visual Encoder model in LLM's ability.
    \item Application of Retrieval-Augmented Generation (RAG) within GPSM4K to enrich multimodal understanding in geometry problem-solving, leveraging a vector database to retrieve relevant question-answer pairs and image descriptions during inference 
\end{itemize}

The remaining of this paper is organized as follows: Section 2 discusses the related works, including existing datasets and methods for geometry problem-solving in Large Vision Language Models (LVLMs). Section 3 presents the GPSM4K dataset, detailing the data collection process, structure, and augmentation techniques. Section 4 describes the experimental setup, including model training and evaluation methodologies. Section 5 showcases the results of our evaluations, highlighting the performance improvements achieved through various techniques such as image captioning and Retrieval-Augmented Generation (RAG). Section 6 provides an in-depth discussion of the implications of our findings, potential limitations, and future research directions. Finally, Section 7 concludes the paper by summarizing our contributions and key takeaways.


\begin{figure*}[h!]
    \centering
    \includegraphics[width=\textwidth]{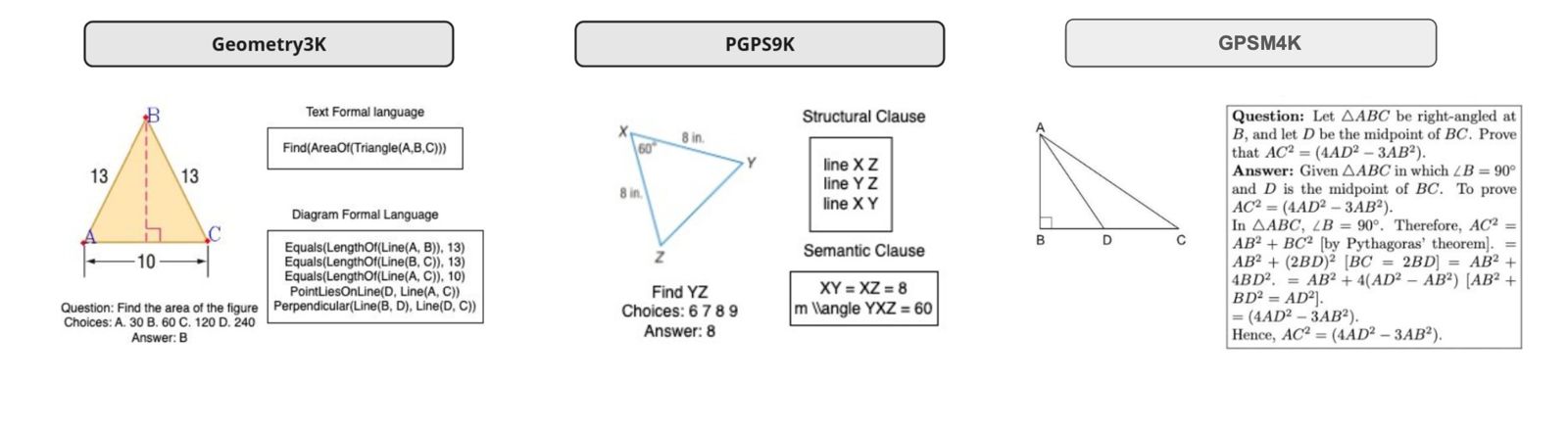}
    \caption{Comparison of Geometry3K, PGPS9K, and our own GPSM4K datasets, showcasing  geometric problems and their representations of complexity in solutions.}
    \label{fig:dataset_Comparison}
\end{figure*}

\section{Related Works}
\subsection{Geometry Multimodal Datasets}
\subsubsection{PGPS9K}
PGPS9k is a comprehensive geometry problem dataset featuring 9,022 geometry problems accompanied by corresponding diagrams. Unlike prior datasets, PGPS9K uniquely provides both diagram annotations and solution programs\cite{zhang2023multi}. Curated from five popular textbooks spanning grades 6-12, it comprises 30 problem types, covering a wide spectrum of plane geometry problems. Noteworthy properties include theorem-based problem-solving, a diagram-centric approach with over 90\% dependency, abstract geometric representation, fine-grained variations in problem formulations, and condition redundancy mitigation.

\subsubsection{Geoemtry3K} 
Geometry3k is a large-scale geometry problem benchmark sourced from two prominent high school textbooks and online digital libraries\cite{lu2021inter}. It encompasses challenging questions like irregular quadrilaterals and polygons and also presents questions with numerous unknown variables and operator types, often necessitating equation solving. Unlike some datasets, Geometry3k lacks annotated theorem application sequences due to resource-intensive annotation requirements.

\subsubsection{Geo170K}
An innovative creation by Gao et al~\cite{gao2023g}, leverages existing datasets through text-only LLMs to synthesize geometric visual-text data. This multi-modal dataset comprises approximately 60,000 geometric image-caption pairs and over 110,000 question-answer pairs, significantly surpassing its predecessors such as GeoQA++. Its expansive coverage and scale, 28 times larger than GeoQA++, enrich the landscape of geometric problem datasets, facilitating broader research and analysis in the domain.

\subsection{Multimodal LLMs }
\subsubsection{LLaVA}
Large Language and Vision Assistant (LLaVA) is a multimodal model designed to be a general-purpose visual assistant. Integrating the capabilities of large language models like GPT-4 with vision encoders such as CLIP \cite{radford2021learning}, LLaVA is uniquely trained to process and act upon multimodal instructions. Unlike current models that predominantly use language for image description, LLaVA enhances interactivity and adaptability by utilizing both visual and textual inputs to execute tasks. Liu et al.~\cite{liu2024visual} present this innovative approach, which transforms image-text pairs into functional instruction-following data.

\subsubsection{GLLaVA}
The authors propose G-LLaVA, a model that leverages the unique characteristics of geometric problems and the capabilities of textual LLMs to enhance understanding and problem-solving in geometry. Current Multimodal Large Language Models (MLLMs) struggle with comprehending basic geometric elements and their relationships, leading to inaccuracies in solving geometric problems. Existing datasets are small and lack comprehensive descriptions of geometric images, which limits the models' ability to understand and solve geometric problems effectively. The authors created a new dataset called Geo170K, consisting of over 170,000 geometric image-caption pairs and question-answer pairs. The GLLaVA model builds on the LLaVA architecture, incorporating LLAMA-2 \cite{touvron2023llama} for language understanding and a pre-trained Vision Transformer (ViT) \cite{radford2021learning} as the image encoder. This setup is enhanced by a projection layer to align the visual features with the LLM’s text dimensionality. The training model involves two key phases: geometric visual-language alignment and geometric instruction tuning, utilizing standard language modeling loss. These phases are designed to enhance the model's capability to interpret geometric data and understand related textual instructions, aiming for superior performance in geometric problem-solving tasks.

\subsection{Retrieval Augmentated Generation}
Large language models equipped with retrieval-augmented generation (RAG) represent a burgeoning field aimed at enhancing answering capabilities by leveraging external knowledge bases. Although the application of RAG with language-only models has been extensively explored, its adaptation into multimodal vision-language
models remain nascent. Going beyond mere answer generation, the primary goal of multimodal RAG is to cultivate the models’ ability to reason in response to relevant queries. Retrieval-augmented generation (RAG) has rapidly emerged as a cornerstone in the development of large language models (LLMs), enabling them to enhance their capabilities by leveraging external knowledge bases \cite{wei2023enhancing},\cite{jiang2023active},\cite{chen2024benchmarking}. Integrating LLMs with RAG has found its most impactful application within language-centric models, where the dynamic interplay between retrieved content and answer generation significantly elevates the quality and relevance of responses \cite{chang2023survey},\cite{zhao2023explainability}. While early works have demonstrated that incorporating directly retrieved information into language models can improve the quality of the generated content \cite{izacard2023atlas}, subsequent developments have involved refinement and mitigated the potential noise associated with the raw retrieval results \cite{asai2023self} \cite{xu2023recomp} \cite{yu2023chain}, thus ensuring that the content generated is not only accurate but also contextually enriched. Although the integration of RAG with large language models has been extensively explored, its
adaptation to multimodal scenarios that encompass both visual and textual inputs remains relatively nascent \cite{yasunaga2022retrieval},\cite{yan2024corrective},\cite{chen2022murag}. The fundamental limitation is critical:
models trained predominantly on textual data struggle to effectively capture the nuanced complexity of visual information, leading to significant gaps in the model’s ability to accurately interpret and reason about visual content. In-context learning (ICL) has revolutionized the functionality of LLMs, enabling
them to adapt to new tasks by leveraging a few contextual examples provided directly within their input. Expanding ICL into multimodal tasks represents a significant advancement, tackling the complex challenge of integrating textual and visual data. The extension of ICL into multimodal tasks represents
a significant leap forward, addressing the inherent complexity of integrating textual and non-textual data. MM-Retrieval \cite{liu2023retrieval} is a concurrent work that introduces a retrieval-augmented multi-modal CoT reasoning approach

\subsection{Small LLMs vs Large LLMs}
Large Language Models (LLMs), such as GPT-4, feature extensive and complex architectures with deep neural networks comprising billions of parameters. This allows them to excel in language understanding and generation. In contrast, Small Language Models (SLMs) are designed with fewer parameters, making them more efficient but somewhat limited in language processing capabilities compared to LLMs. LLMs have shown impressive abilities to tackle a wide range of tasks without needing specific fine-tuning on task-specific datasets. However, deploying LLMs in practical applications is challenging due to the significant computing resources required. The instruction-following capabilities of LLMs have enabled them to perform exceptionally well in zero-shot scenarios \cite{tahmid2023systematic}, \cite{bang2023multitask} increasing their use in solving real-world problems. Nevertheless, experimental results indicate that most smaller LLMs, even after fine-tuning, do not surpass the performance of larger zero-shot LLMs. For example, GPT-4 is currently considered the top-performing LLM across various evaluation benchmarks. These closed-source LLMs, available only through APIs, also raise potential privacy concerns. To address these issues, various open-source LLMs have been developed \cite{touvron2023llama},\cite{jiang2023mistral}. The main advantages of using open-source LLMs include (i) the ability for in-house deployment, (ii) the potential for fine-tuning to achieve performance comparable to larger closed-source LLMs, and (iii) the same inference cost for both zero-shot and fine-tuned versions. Thus, open-source LLMs present a promising alternative to overcome the limitations of closed-source LLMs.

\subsection{Geometry Problem Solving}
The Geometry problem reasoning is a challenging visual mathematical reasoning problem. Initial efforts by \cite{seo2015solving} , \cite{sachan2017textbooks} . \cite{sachan2017learning} focused on creating datasets manually. More recent approaches have introduced improved methods and datasets, such as Geometry3K \cite{lu2021inter}, GeoQA \cite{chen2021geoqa}, GeoQA+ \cite{cao2022augmented}, aiming to enhance both performance and explainability. Nevertheless, traditional models' performance in this domain has not reached the level seen in other areas of mathematical problem solving, especially when compared to approaches that use large language models for solving math word problems \cite{cobbe2021training} \cite{wei2022chain} \cite{gou2023tora}.

\subsection{Data Generation via LLM}
Bootstrapping data from pretrained models has long been a focus of research. \cite{ye2022zerogen} \cite{meng2022generating} generate training data using pretrained language models like GPT-2 for classification tasks. \cite{gao2022self} employs influence functions to select in-context examples for data generation. Recently, with the advent of powerful LLMs such as ChatGPT, automatic data generation has become more widespread. A series of recent works use ChatGPT-generated data for instruction tuning, including \cite{wang2022self} \cite{peng2023instruction} \cite{taori2023stanford},\cite{liu2024visual}.

\subsection {Need for Dataset Creation}
Current benchmark datasets for geometry often fall short in their complexity, particularly with numerical problems and theorem-based questions. Although they include basic geometric concepts and visual elements, they overlook the more detailed numerical computations and theorem-based queries typical in high school geometry courses. Additionally, these datasets frequently do not provide detailed, step-by-step solutions that are vital for understanding how to solve problems and learn from errors. Consequently, there is a clear need for a specialized dataset designed specifically for geometry, one that covers a wider spectrum of complexities including advanced numerical tasks, theorem-based questions, and comprehensive solutions.

\begin{figure*}[ht]
    \centering
    \includegraphics[width=\textwidth]{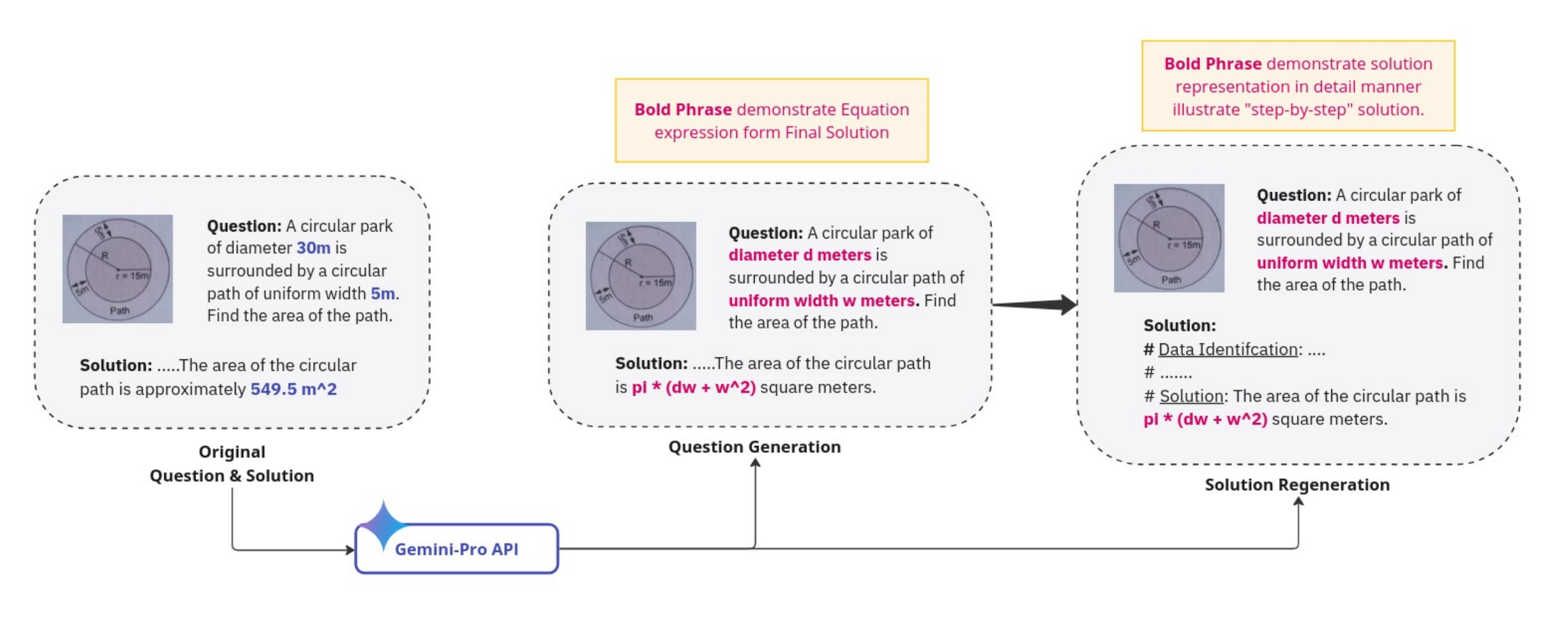}
    \caption{Dataset augmentation Pipeline using the Gemini-Pro API. The process starts with the original question and solution, generates new questions with varied parameters, and regenerates detailed step-by-step solutions with equation expressions highlighted in bold. The augmentation enhances the dataset by introducing variability and detailed solutions.}
    \label{fig:augmentation_pipeline}
\end{figure*} 

\section{Methodology}
Our study developed "GPSM4K" a comprehensive geometry dataset formed from Grades 6 through 12, sourced from textbooks aligned with State Board and NCERT standards. Unlike traditional datasets that primarily consist of multiple-choice questions, GPSM4K features Numerical Answer Questions and Theorem Proving Questions, enhancing its educational value and breadth. The dataset, formatted in JSON, includes chapter titles, questions, answers, associated imagery, and educational levels, covering a broad spectrum of geometric topics. Each question is accompanied by two solutions: the original textbook solution and a restructured version via Gemini Pro vision, which breaks down solutions into fundamental components, fostering a methodical approach to problem-solving and enhancing the interpretability of computational models.

\subsection{Dataset Extraction}

\begin{figure}[h!]
    \centering
    \includegraphics[width=8cm]{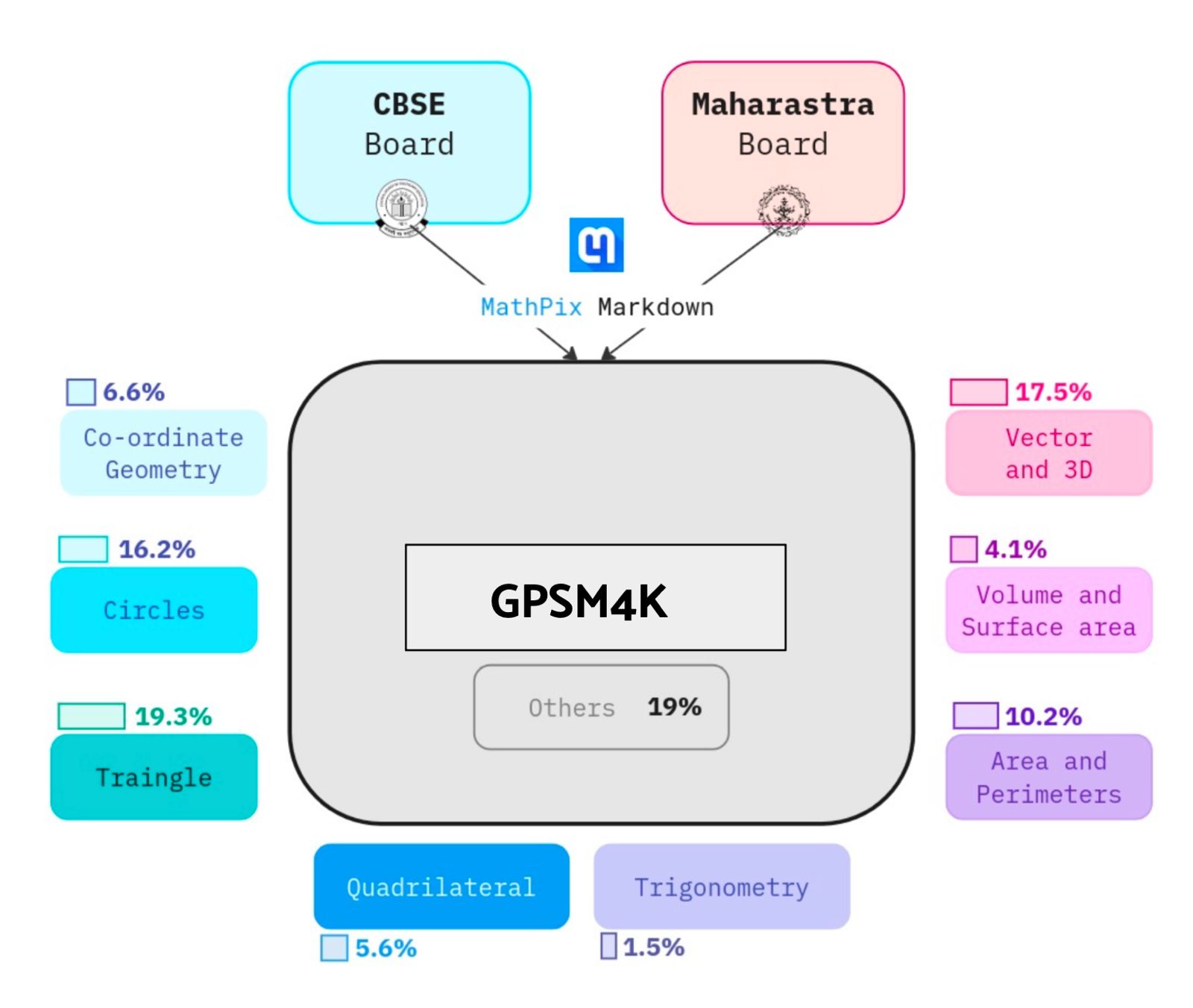}
    \caption{Data Extraction and topic-wise distribution}
    \label{fig:dataset_distribution}
\end{figure}

This was meticulously executed by leveraging resources from high-school textbooks and previous years' question papers across various Indian education boards, including CBSE (Central Board of Secondary Education), and the Maharashtra Board. These educational materials were accessed online in PDF format and processed through Mathpix, a tool adept at converting PDF documents into Mathpix Markdown (MMD) and subsequently into LaTeX documents. This conversion facilitated the extraction of geometric question-answer pairs, incorporating the richness of LaTeX's features with the simplicity of Markdown. Our focus was on questions related to geometry that were accompanied by images. The team of human experts extracted these questions, along with their corresponding solutions, images, standard (grade), and specific topics within geometry. This rigorous process ensured the creation of a comprehensive and relevant dataset for our research. Figure \ref{fig:dataset_distribution} shows chapter chapter-wise distribution of our dataset across 9 distinct geometric concepts and the statistics can be seen in Table \ref{tab:dataset_summary}.

\begin{table*}[h]
\caption{GPSM4K Finetuning Dataset Statistics}
\label{tab:dataset_summary}
\centering

\begin{tabular}{c|c|c|c|c}
\toprule
\textbf{Difficulty } & \textbf{Question Type} & \textbf{Original} & \textbf{Augmented} & \textbf{Total} \\
& & \textbf{Samples} & \textbf{Samples} & \textbf{Samples} \\
\midrule
Easy & Numericals & 250 & 500 & 750 \\ \\
Medium & Numericals & 700 & 1400 & 2100 \\ \\
Hard & Numericals & 140 & 280 & 420 \\ \\ \hline \\
Easy & Theorems & 109 & 218 & 327 \\ \\
Medium & Theorems & 300 & 600 & 900 \\ \\
Hard & Theorems & 70 & 140 & 210 \\ \\
\midrule
Total Samples & & \textbf{1480} & \textbf{2960} & \textbf{4440} \\ \hline
\bottomrule
\end{tabular}
\end{table*}


\begin{table}[h]
\caption{GPSM4K Inference Dataset Statistics}
\label{tab:dataset_summary_2}
\centering

\begin{tabular}{c|c|cc}
\toprule
\textbf{Difficulty } & \textbf{Question Type} & \textbf{Original Samples}  \\ \midrule
Easy & Numericals & 70 \\ \\
Medium & Numericals & 56 \\ \\
Hard & Numericals & 34 \\ \\ \hline \\
Easy & Theorems & 18  \\ \\
Medium & Theorems & 12   \\ \\
Hard & Theorems & 10   \\ \\
\midrule
Total Inference Samples & & \textbf{200} \\ \hline 
\bottomrule
\end{tabular}
\end{table}

\subsection{Dataset Augmentation}

\subsubsection{\textbf{Diagram Description Generation}} 
In this phase, the Gemini Vision Pro API was utilized to automatically generate descriptions of diagrams, playing a vital role in enriching the dataset with crucial visual information. This step was essential for building a comprehensive multi-modal question-answering system.

The process involved feeding the API with a geometric question-answer pair alongside the relevant diagram. The prompt was carefully crafted to ensure that the descriptions generated were focused solely on the visual characteristics of the diagram, avoiding any overlap with the content of the question or the answer. The goal was to produce concise descriptions that were directly aligned with the key visual features of the diagram. In addition to the prompt, the geometric question, answer, and the diagram were provided, allowing the model to generate a precise description that accurately highlighted the diagram’s essential geometric attributes.

\subsubsection{\textbf{Question Generation}}
By leveraging Gemini Pro Vision \cite{team2023gemini}, a multimodal large language model (LLM), new question-answer pairs were systematically generated from textbook-derived diagrams. For each original diagram-QA pair, two new question-answer pairs were generated. This process considered both the visual details embedded in the diagrams and the textual context of the existing questions and answers. Through this augmentation approach, the dataset was significantly expanded by introducing variations in how the questions were framed while retaining semantic consistency with the source material. The aim was to enable multimodal models to handle a wider array of question formulations, improving their ability to understand and respond to diverse types of inputs.

\subsubsection{\textbf{Solution Regeneration}}
Following the question generation, solutions were regenerated using Gemini Pro, a text-based LLM. A structured prompt was employed to deconstruct and reconstruct solutions methodically. This process involved breaking the problem into its core components, highlighting the relevant concepts and theories, and developing systematic methodologies to solve the problem. The computations were performed with precision, leading to the presentation of a refined solution. The purpose of this regeneration step was to promote a more profound understanding of mathematical concepts and problem-solving techniques in multimodal models, thereby enabling them to handle complex queries with greater effectiveness. Through this systematic regeneration approach, the multimodal models were trained to identify underlying patterns and relationships, thereby enhancing their comprehension and response capabilities. The entire data augmentation pipeline is illustrated in \ref{fig:augmentation_pipeline}, showcasing the steps as described above.

\begin{figure}[h]
\begin{tcolorbox}
{\scriptsize
\begin{verbatim}
# Question : {Geometry Questions}
# Image : {Correesponding Image} 

# Data Identification: "[List the key values and information provided 
in the question. 
Format this section as a JSON string.]"

# Problem Analysis: "[Break down the problem into its fundamental components. 
Present this analysis as a JSON string.]"

# Theoretical Framework: "[Outline the relevant concepts, theories, or 
principles that apply to the problem. Format this explanation as a 
JSON string.]"

# Methodology Development: "[Develop the formulae, algorithms, or theorems 
necessary to solve the problem, if applicable. Describe these methodologies 
in a JSON string format.]"

# Computation: "[Perform the necessary calculations or logical steps to 
arrive at the solution. Present these computations as a JSON 
string format.]"

# Solution: "[Present the final answer or solution to the problem in a JSON 
string format. Ensure clarity and conciseness in the explanation.]"
\end{verbatim}
}
\end{tcolorbox}
\caption{Prompt used for Solution Regeneration}
\label{fig:regen_prompt}
\end{figure}


 \begin{center}
    \includegraphics[width=0.3\textwidth]{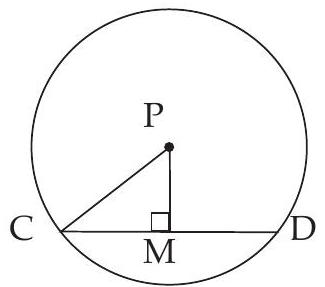}
\end{center}

\textbf{Original Q\&A Pair}
\medskip
\hrule
\medskip
\textbf{Question:}
The diameter of a circle is $26 \mathrm{~cm}$ and the length of the chord of a circle is $24 \mathrm{~cm}$. Find the distance of the chord from the center.

\textbf{Answer:}
Given:\\
(i) A circle with centre 'P' and diameter $26 \mathrm{~cm}$. \\
(ii) Length of chord $\mathrm{CD}=24 \mathrm{~cm}$ \\
(iii) seg $\mathrm{PM} \perp$ chord $\mathrm{CD}$, $\mathrm{C}-\mathrm{M}-\mathrm{D}$ \\

Diameter of the circle $=26 \mathrm{~cm}$ \\
...(Given) \\
Radius $=\frac{\text { Diameter }}{2}=\frac{26}{2}$ \\
$\therefore \quad$ Radius of the circle $=13 \mathrm{~cm}$ \\
$\therefore \quad \mathrm{PC}=13 \mathrm{~cm}$ \\
Seg $\mathrm{PM} \perp$ chord $\mathrm{CD}$ \\
...(Given) \\
$\therefore \quad \mathrm{CM}=\frac{1}{2} \mathrm{CD}$...(Perpendicular drawn from the centre of the circle to the chord bisects the chord.) \\
$$
\mathrm{CM}=\frac{1}{2} \times 24=12 \mathrm{~cm} \\
$$
In $\triangle \mathrm{PMC}, \angle \mathrm{PMC}=90^{\circ}$ \\
...(Given) \\
$\therefore \quad \mathrm{PC}^{2}=\mathrm{PM}^{2}+\mathrm{CM}^{2} \ldots$ (Pythagoras theorem) \\
$\therefore \quad 13^{2}=\mathrm{PM}^{2}+12^{2}$ \\
$\therefore \quad 169-144=\mathrm{PM}^{2}$ \\
$\therefore \quad \mathrm{PM}^{2}=25$ \\
$\therefore \quad \mathrm{PM}=5 \mathrm{~cm}$ \\
...(Taking square roots)

    
    \bigskip
    \hrule
    \bigskip

    \textbf{Diagram Description Generation}:
    \medskip
    \hrule
    \medskip
A circle with center P and a diameter of 26 cm. A chord CD of length 24 cm is drawn in the circle. PM is perpendicular to CD and M is the mid-point of CD.
    
    \bigskip
    \hrule
    \bigskip


    \textbf{Question Augmentation}
    \medskip
    \hrule
    \medskip
    \textbf{Question 1:} \\
The diameter of a circle is $13 \, \mathrm{cm}$, and the length of a chord of the circle is $12 \, \mathrm{cm}$. Find the distance of the chord from the center. \\

\textbf{Answer 1:} \\
Let's denote the radius of the circle as $r$.

Given that the diameter of the circle is $13 \, \mathrm{cm}$, so the radius, $r$, is half of the diameter, i.e., $r = \frac{13}{2} = 6.5 \, \mathrm{cm}$.

Since $PM$ is perpendicular to $CD$, it bisects $CD$, so $CM = \frac{1}{2} CD$.

\[
CM = \frac{1}{2} \times 12 = 6 \, \mathrm{cm}
\]

In $\triangle PMC$, as $\angle PMC = 90^\circ$, by Pythagoras' theorem:

\[
PC^2 = PM^2 + CM^2
\]

Substituting the values, we get:

\[
6.5^2 = PM^2 + 6^2
\]
\[
42.25 = PM^2 + 36
\]
\[
PM^2 = 42.25 - 36 = 6.25
\]
\[
PM = \sqrt{6.25} = 2.5 \, \mathrm{cm}
\]

Thus, the distance of the chord from the center is $2.5 \, \mathrm{cm}$.

\section{Experimentation}

In this study, we perform a series of experiments to evaluate the performance of Large Vision Language Models (LVLMs) utilizing our newly developed GPSM4K dataset. These experiments were conducted to assess the models' abilities across a range of scenarios, including zero-shot inference, fine-tuning, and the influence of multimodal enhancements. We experimented with several configurations of the LLaVA models (versions 1.5 and 1.6), GLLaVA models, and Gemini Pro Vision, comparing their performance on multiple datasets such as GPSM4K, PGPS9K, and Geometry3K. For the LLaVA models, we employed configurations using Vicuna 7/13B \cite{zheng2024judging} and Mistral 7B \cite{jiang2023mistral}, among others. Furthermore, we explored the impact of techniques like image captioning and Retrieval-Augmented Generation (RAG) in improving model performance. All results are reported based on experiments conducted on a test set of 150 geometry problems.

The experimental design aimed to offer a thorough evaluation of the models' generalization capabilities, the role of visual encoders, and the advantages of incorporating multimodal data into geometric problem-solving tasks. Detailed descriptions of the methodologies and the outcomes for each experiment are presented in the subsequent sections.

\subsection{Zero-shot Inference Analysis :}


In the zero-shot inference phase, our objective is to assess the models' generalization abilities. We employ two distinct models, LLaVA \cite{liu2024visual} and G-LLaVA \cite{gao2023g}, for this purpose. These models are exposed to geometry questions from our GPSM4K dataset, which they have not encountered during their pretraining phase. The aim is to evaluate their ability to apply overarching knowledge and deductive reasoning skills to solve specific and new geometry problems without any prior exposure. This approach is crucial in understanding the extent to which the models can infer and solve problems outside their training scope, reflecting their real-world applicability and generalization capabilities.

The results, presented in Table \ref{tab:accuracy}
, reveal that GPT-4 achieved the highest accuracy across all datasets, demonstrating superior generalization capabilities. Specifically, GPT-4 achieved 60\% accuracy on GPSM4K, 36\% on PGPS9K, and 51\% on Geometry3K. Gemini-Pro also performed well, with 44\% accuracy on GPSM4K, 30\% on PGPS9K, and 47\% on Geometry3K. Among the LLaVA models, LLaVA-1.5-13B and LLaVA-1.6-13B showed comparatively better performance. G-LLaVA-7B also exhibited strong generalization with 18\% accuracy on GPSM4K, 10\% on PGPS9K, and 21\% on Geometry3K. These results highlight the effectiveness of GPT-4 and Gemini-Pro in zero-shot inference scenarios, while also indicating room for improvement in the GLLaVA-based models.
The superior performance of Gemini Pro and GPT-4 in zero-shot inference is largely due to their extensive parameter sizes and advanced training on a vast range of data, enabling them to generalize better to new problems.
\
\begin{table}[h!] 
\centering
\caption{Comparison of the Accuracy (\%) on different models with Zero-shot inference on GPSM4K  (Ours) dataset and Benchmark Dataset: PGPS9K and Geometry3K}

\label{tab:accuracy}

\begin{tabular}{l|c|c|c}
\toprule
\textbf{Model} & \textbf{GPSM4K} & \textbf{PGPS9K} & \textbf{Geometry3K} \\
\midrule
LLaVA-1.5-7B & 3 & 4 & 1 \\ \\
LLaVA-1.5-13B & 7 & 2 & 14 \\ \\
LLaVA-1.6-7B & 3 & 1 & 11 \\ \\
LLaVA-1.6-13B & 3 & 8 & 7 \\ \\
G-LLaVA-7B  & 18 & 10 & 21 \\ \\
Gemini-Pro  & 44 & 30 & 47 \\ \\
GPT-4  & \textbf{60} & \textbf{36} & \textbf{51} \\
\bottomrule
\end{tabular}
\end{table}

\subsection{Experiment 1
}
\textbf{\textit{"Why do we need different datasets for Geometry Problems ?"
}} \\
In this experiment, we selected the PGPS9K dataset, consisting of plane geometry problems, featuring over 9,000 problems with fine-grained diagram annotations and interpretable solution programs, to serve as a benchmark in our study. We fine-tune LLaVA and G-LLaVA variants on the PGPS9K dataset and our custom-developed GPSM4K  dataset to evaluate their performance on our test set of 150 high-school-level geometry problems. Models are fine-tuned with a learning rate of 3e$^{-5}$, a batch size of 4 per GPU, and a total of 2 epochs. This process utilized an NVIDIA DGX server with A-100 GPUs, each equipped with 40 GB of RAM. This experiment evaluates the effectiveness of models in addressing high school-level geometric problems that demand step-by-step solutions involving the application of different theorem knowledge and numerical reasoning. 
The results presented in Table \ref{tab:
} indicate that models fine-tuned on the GPSM4K dataset generally outperformed those fine-tuned on the PGPS9K dataset. For example, the base LLaVA 1.6 models with Vicuna and Mistral configurations showed significant improvements in accuracy when trained on our dataset. Notably, GPT-4 achieved the highest accuracy of 56.66\%, followed by Gemini with 35.33\%, demonstrating the effectiveness of these models in handling the GPSM4K test set.

 The superior performance of models trained on the GPSM4K dataset can be attributed to its detailed step-by-step solutions, which provide a more comprehensive understanding of the problem-solving process. In contrast, the PGPS9K dataset primarily features multiple-choice questions (MCQs) with corresponding diagrams, which may not offer the same depth of reasoning required for complex problem-solving. 

\begin{table}[ht]
\caption{Performance of LVLMs on GPSM4K test set}
\label{tab:experiment_1}

\begin{tabular}{p{4cm}|l|c}
\toprule
\textbf{Models} & \textbf{Trained on} &\textbf{Accuracy (\%) on } \\
& &  \textbf{GPSM4K test set} \\  
\midrule
Base LLaVA 1.6 + Vicuna 7B & PGPS9K & 8\\ \\
Base LLaVA 1.6 + Vicuna 13B & PGPS9K & 8.66\\ \\
Base LLaVA 1.6 + Mistral 7B & PGPS9K  & 7.33\\ \\
Base LLaVA 1.6 + Vicuna 7B & Ours  & 22.66 \\ \\
Base LLaVA 1.6 + Vicuna 13B & Ours & 24 \\ \\
Base LLaVA 1.6 + Mistral 7B & Ours  & 24.66 \\ \\
Base LLaVA 1.6 + Vicuna 34B & -  & 27.33 \\ \\
Base LLaVA3 8B  & -  & 22.6 \\   \\
GLLaVA 7B + LLaMa 1.5 & GeoQA++ & 10.66\\ \\
GLLaVA 13B + LLaMa 1.5 & GeoQA++ & 22.66 \\ \\
GLLaVA 7B + LLaMa 1.5 & Ours & 17.33 \\ \\
GLLaVA 13B + LLaMa 1.5 & Ours & 25.33\\ \\
Gemini & -  & 35.33\\ \\
GPT-4 & - & \textbf{56.66}\\ \\
\bottomrule
\end{tabular}
\end{table}

\subsection{Experiment 2}

\textbf{\textit{"Does the Visual Encoder model restrict LLMs' ability to understand math, especially geometry, due to its inability to decode geometric figures accurately, lacking standard features found in common image datasets?"}} \\

The effectiveness of combining Visual Encoder models with Language Models (LMs) for understanding mathematics, particularly geometry, deserves careful examination. Modern vision models like LLaVa\cite{liu2024visual}, which are usually trained on image datasets like COCO or ImageNet, may struggle with decoding geometric figures because these datasets typically lack the specialized features found in geometric imagery. This shortfall can make it difficult for LMs that depend on visual data to correctly interpret relevant information and solve intricate geometry problems. Therefore, it's crucial to assess how Visual Encoder models affect LMs' ability to perform mathematical reasoning tasks, especially those involving geometry. 

In this experiment, we utilize images from our test set of 150 examples, leveraging models such as GIT\cite{wang2022git}, BLIP\cite{li2022blip}, its successor BLIP-2 \cite{li2023blip}, LLaVa\cite{liu2024visual}, Vision Transformer\cite{dosovitskiy2020image}  and Gemini Pro to generate captions. These are then compared with human-generated captions provided by our subject matter experts using the Jaccard index and Cosine similarity metrics.

We chose Jaccard and Cosine similarities over traditional metrics like BLEU \cite{papineni2002bleu} and ROUGE\cite{lin2004rouge} which focus heavily on exact n-gram matches, which can be too rigid for image captioning where diverse but equally valid captions are possible. Cosine similarity handles synonyms and paraphrasing better than n-gram-based metrics, making it particularly useful in image captioning where different words can accurately describe the same image. Additionally, both Jaccard and Cosine similarities are computationally less intensive compared to metrics involving deep learning models like BERTScore, making them more practical for large-scale evaluations or scenarios with limited computational resources

\begin{figure*}[ht]
    \centering
    \includegraphics[width=\textwidth]{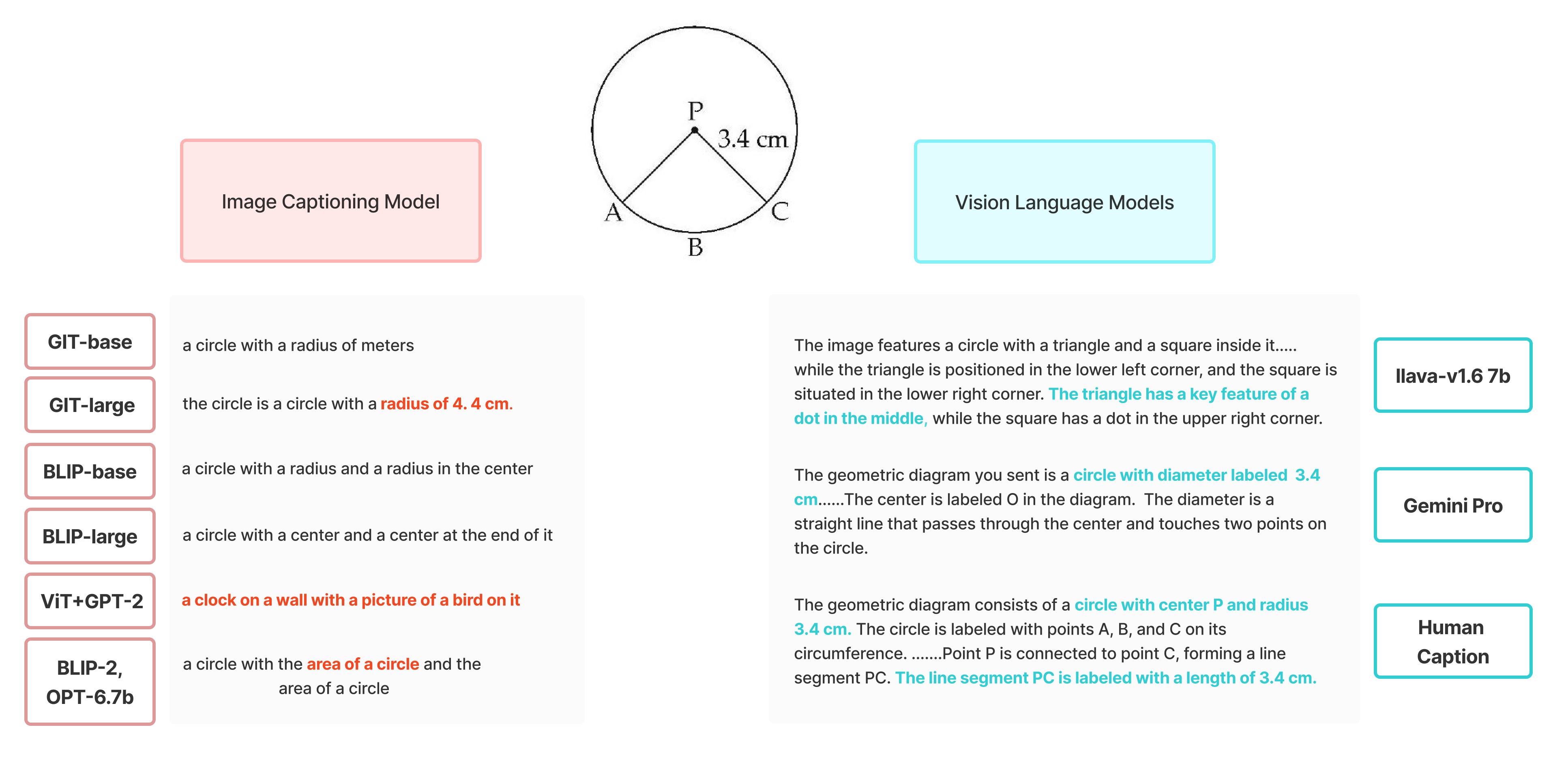}
    \caption{Image captioning example}
    \label{fig:image_caption}
\end{figure*}

Figure \ref{fig:image_caption} shows captions generated by various models on a sample geometry diagram.
The table \ref{tab:captions} clearly illustrates that Gemini Pro, which likely incorporates more sophisticated mechanisms for interpreting visual data alongside textual information, outperforms other models in both the Jaccard index and cosine similarity metrics.
\begin{table}[ht]
\centering
\caption{Image Captioning Result}
\label{tab:captions}

\begin{tabular}{l|l|l}
\toprule
\textbf{Models} & \textbf{Mean / Median } & \textbf{Mean / Median} \\
& \textbf{Jaccard} & \textbf{Cosine similarity} \\ \midrule
GIT-base  & 0.08 / 0.07 & 0.25 / 0.24 \\ \\
GIT-large & 0.09 / 0.09 & 0.28 / 0.27 \\ \\
BLIP-base & 0.11 / 0.10 & 0.23 / 0.22\\ \\
BLIP-large & 0.14 / 0.13 & 0.25 / 0.23\\ \\
ViT+GPT-2 & 0.05 / 0.05 & 0.06 / 0.06 \\ \\
LLaVa v1.6 7b & 0.12 / 0.12 & 0.43 / 0.42\\ \\
BLIP-2 OPT-6.7b & 0.12 / 0.12 & 0.33 / 0.32 \\ \\
{\textbf{Gemini Pro}} & {\textbf{0.36 / 0.28}} & {\textbf{0.65 / 0.63}} \\ \bottomrule
\end{tabular}
\end{table}

\subsection{Experiment 3}

 
\textbf{\textit{"Can adding image captions to multimodal geometry figures improve LLMs' ability to solve complex geometry problems and theorems?"}}\\

 In this experiment, we explored how adding image captions affects the performance of LVLMs in solving geometry problems. Based on the results of Experiment 2 table \ref{tab: experiment_4}, where Gemini Pro produced better captions compared to other models based on the Jaccard index and Cosine similarity metrics, we chose Gemini Pro and human-generated captions on the test set for this experiment. We then integrated these captions with the question prompts for inference purposes and fed them to the GLLAVA 7B and 13B models, both of which were finetuned on our GPSM4K dataset.
The results, summarized in Table \ref{tab
}, show a significant improvement in model performance with the addition of captions, highlighting the importance of combining textual and visual information.
Interestingly, the LLaMa3 8B model, which used only text without any image captions, performed exceptionally well with an accuracy of 28.66\%. This suggests that when provided with high-quality text descriptions, the model can effectively solve geometry problems, demonstrating the critical role of detailed and accurate captions.
Also, the accuracy of GLLaVA 7B and 13B models increased with the addition of captions, with human-generated captions providing a greater boost than those generated by Gemini Pro.

 \begin{table}[htbp]
\centering
\caption{Model performance on GPSM4K test after adding Image captions}
\begin{tabular}{l|c}
\toprule
\textbf{Model Finetuned on GPSM4K + Captions} & \textbf{Accuracy (\%) on } \\
& \textbf{GPSM4K test set} \\
\midrule
GLLaVA 7B FT + Gemini   & 21.33 \\ \\
GLLaVa 13B FT + Gemini   & 22.66 \\ \\
GLLaVA 7B FT + Human  & 24.66 \\ \\
GLLaVa 13B FT + Human  & 26.66 \\ \\
\bottomrule
\end{tabular}
\label{tab: experiment_3}
\end{table}

\subsection{Experiment 4}
\begin{figure*}[ht]
    \centering
    \includegraphics[width=\textwidth]{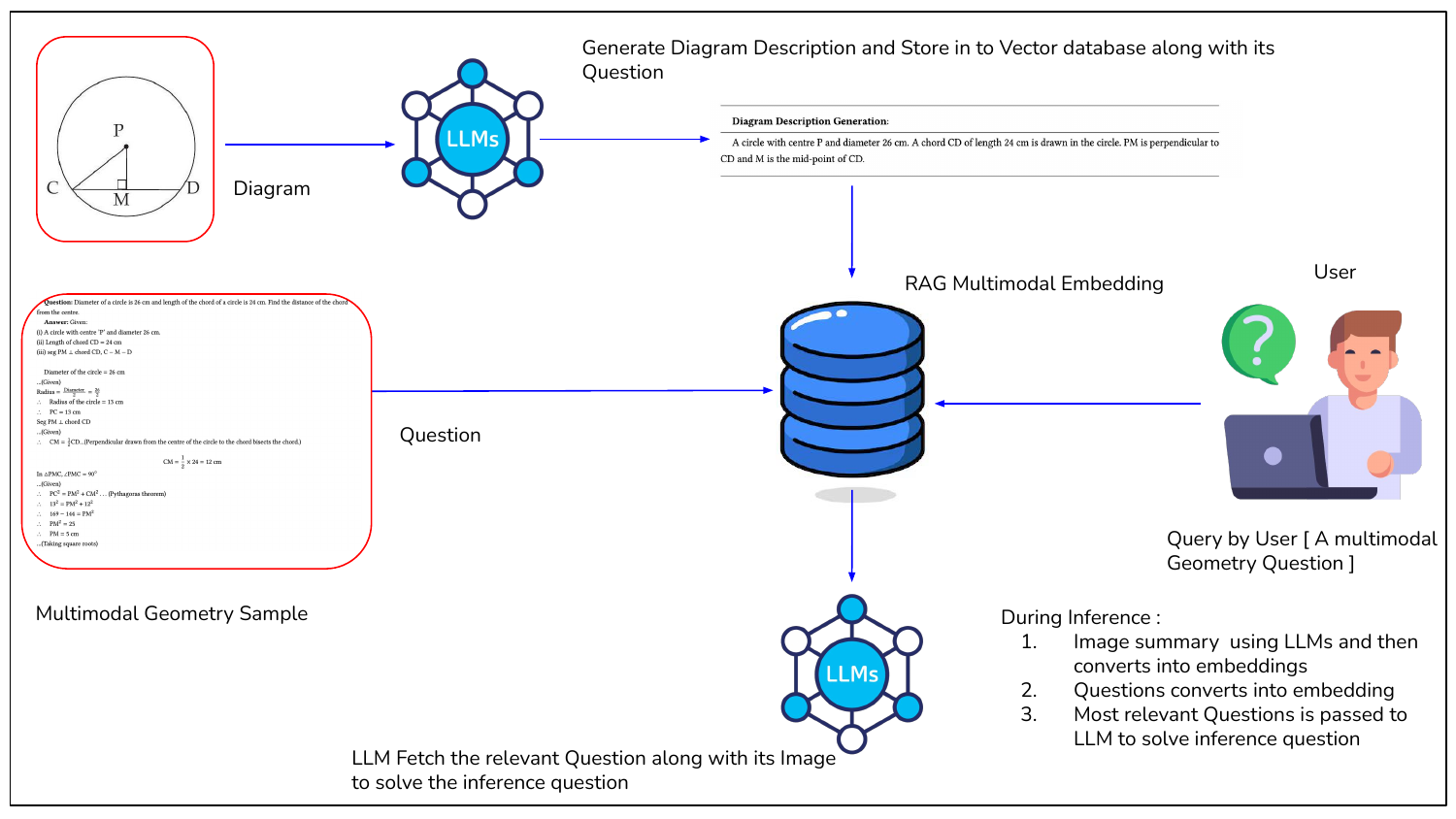}
    \caption{Multimodal RAG tackles this challenge by combining the strengths of images and descriptions. It analyzes diagrams to understand the problem and uses text descriptions to find the solution.In 1-shot multimodal RAG, a single problem-solution example is used to find similar problems from a vast dataset. These retrieved examples then help solve new inference questions, showcasing the model's ability to generalize its knowledge}
    \label{fig:mmrag}
\end{figure*}

\textbf{\textit{"Can Multi-modal RAG improve performance of models 
on Geometry problems? "}}\\
We develop a comprehensive multimodal RAG framework \ref{fig:mmrag},  which seamlessly integrates multimodal retrieval capabilities with in-context learning (ICL). This framework begins by employing a retrieval module to fetch the most pertinent question-answer pairs, which is multimodal. It then integrates these elements into the model’s reasoning process, guiding it through the provided rationales associated with each retrieved item. Following this retrieval phase, the model autonomously learns coherent rationales that reflect a deep and meaningful engagement with the given problem. Retrieval-augmented generation (RAG) enhances language models by retrieving relevant data from a large corpus and feeding this data into a generative model to inform its output. It is particularly effective in generating knowledgeable and context-aware answers by combining the strengths of both retrieval and generative mechanisms.
For our experiments, we used a vector database to store question-answer pairs along with descriptions of associated images from our training dataset. During the inference process, when presented with a new question, we utilized cosine similarity to retrieve closely related questions from our vector DB. This retrieved information served as context for the model, helping to provide illustrative examples that enrich the model's response. We conducted 1-shot RAG experiments on the Gemini Pro and LLaMa3 models, with the outcomes detailed in Table \ref{tab: experiment_4}\\

 \begin{table}[htbp]
\centering
\caption{Model performance on GPSM4K test with RAG}
\begin{tabular}{p{5cm}|c}
\toprule
\textbf{Model} & \textbf{Accuracy (\%) on } \\
& \textbf{GPSM4K test set} \\
\midrule
Gemini   & 35.33 \\ \\
Gemini + 1-shot RAG   & \textbf{38} \\ \\
LLaMa 3 8B  & 22.6 \\ \\
LLaMa 3 8B + 1-shot RAG  & \textbf{26} \\ \bottomrule
\end{tabular}
\label{tab: experiment_4}
\end{table}

\section{Evaluation}
In our evaluation, we utilized the Gemini Pro API to assess the accuracy of our models. We employed two main methods: final answer evaluation and step-by-step evaluation. The primary metric for evaluation was Top-1 accuracy, which differs from the commonly used Top-10 accuracy in previous studies. Top-1 accuracy considers only the first generated sequence, while Top-10 accuracy considers any of the ten generated sequences as correct if they solve the problem. We also used a majority count approach, repeating the extraction and comparison process three times to reduce the impact of anomalies or errors in single iterations, leading to more reliable and stable measurements of accuracy.

\subsection{\textbf{Final Answer Evalaution}}

The evaluation of the final solution follows a two-step approach using the Gemini Pro API. In the first stage, the initial prompt examines both the ground truth solution and the predicted outcome, extracting the final answer from each of them. Then, in the second stage, the extracted answers are compared; if they match, they are labeled as "Yes", whereas mismatches are marked as "No". We opted to use two distinct prompts rather than a single one because this approach allowed the model to generate more precise results.
\begin{figure}[htbp]
\centering
\begin{tcolorbox}
\begin{alltt}
{\scriptsize
\textbf{
Prompt 1: Extract the final answer from the Ground Truth and the Prediction.

        Instructions:
        1. Read the final answer in the Ground Truth.
        2. Read the final answer in the Prediction.
        3. Return the final answers in a structured format.

        Ground Truth: <ground truth>
        Prediction: <prediction>

        # Output should be in the format:
        # Ground Truth Answer: <extracted answer>
        # Prediction Answer: <extracted answer>

Prompt 2: Objective: Compare the final answers and label the comparison.

        Instructions:
            1. Compare the final answer from the Ground Truth and the Prediction.
            2. Label the comparison as "Yes" if both answers are identical, and 
            "No" if they differ.
        
        Ground Truth Answer: <ground truth>
        Prediction Answer: <prediction>


        # Output should be in the format:
        # Comparison Result: Yes/No
}
}
\end{alltt}
\end{tcolorbox}
\caption{Two Prompts used in Final Solution Extraction and Comparison respectively}
\label{fig:prompts}
\end{figure}

\begin{figure}[htbp]
\centering
\begin{tcolorbox}
\begin{alltt}
{\scriptsize
\textbf{
Prompt: Evaluate the prediction against the ground truth based on mathematical accuracy.
    
    # Provide a label 'Yes' or 'No' based on the below criteria.

    # Criteria:
    # 1. Mathematical Concept and Computation Steps: Are they the same in 
    both prediction and ground truth?
    #    This includes the approach, steps followed, and principles applied.
    # 2. Final Answer: If applicable, does the final answer match both 
    prediction and ground truth?

    # Labeling:
    # - Label 'Yes' if the mathematical concept, computation steps, and final 
    answer (if applicable) are the same.
    # - Label 'No' if there are differences in the concept, steps, or final 
    answer (if applicable).

    # Provide a Reason for your label.

    # Ground Truth: <ground truth>

    # Prediction: <prediction>

    # Label: Yes/No
    
}}
\end{alltt}
\end{tcolorbox}
\caption{Prompt used for step-by-step evaluation}
\label{fig:prompts1}
\end{figure}

\subsection{CoT Step-by-step-evaluation}

Compared to visual question-answering in general scenarios, solving mathematical problems with MLLMs requires nuanced, step-by-step chain-of-thought (CoT) reasoning. This makes the binary ‘Correct’ or ‘Incorrect’ evaluation approach of existing benchmarks inadequate for examining the depth and precision of multi-step reasoning processes. To address this, we propose a CoT evaluation strategy to thoroughly assess mathematical CoT skills in visual contexts, using two prompting phases with Gemini Pro.

In our evaluation, we only prompt Gemini Pro with the MLLM’s output, deliberately omitting the original questions, diagrams, and ground-truth answers. We do use a pre-defined key-step template for each problem to ensure consistency and comparability, but we also allow for adaptive extraction tailored to the unique output of each MLLM when necessary. This is crucial because mathematical problems often allow for diverse solution pathways, and different MLLMs exhibit varying reasoning lengths and styles. A rigid template can be supplemented with adaptive strategies to maintain the accuracy of CoT evaluations.

For the step-by-step evaluation, we employed a detailed prompt as shown in figure  \ref{fig:prompts1} to evaluate the mathematical accuracy of each step in the problem-solving process. This method assesses the alignment between the prediction and the ground truth in terms of mathematical concepts, computation steps, and the final answer. This approach allows for a granular assessment of the models' problem-solving capabilities, ensuring that not only the final answer but also the intermediate steps and reasoning are evaluated for accuracy and correctness.

Gemini Pro is used to evaluate each critical step and provide a comprehensive score. We input the extracted key steps, original questions, diagrams, and ground-truth answers into Gemini Pro. Specifically, Gemini Pro assesses each intermediate step with a binary score of ‘1’ (correct) or ‘0’ (incorrect) and calculates the overall score by aggregating the correctness of the final answer. This CoT evaluation approach also offers a detailed error analysis for each step, providing valuable insights for the development of MLLMs in the field.

\begin{figure*}[h!]
    \centering
    \includegraphics[width=\textwidth]{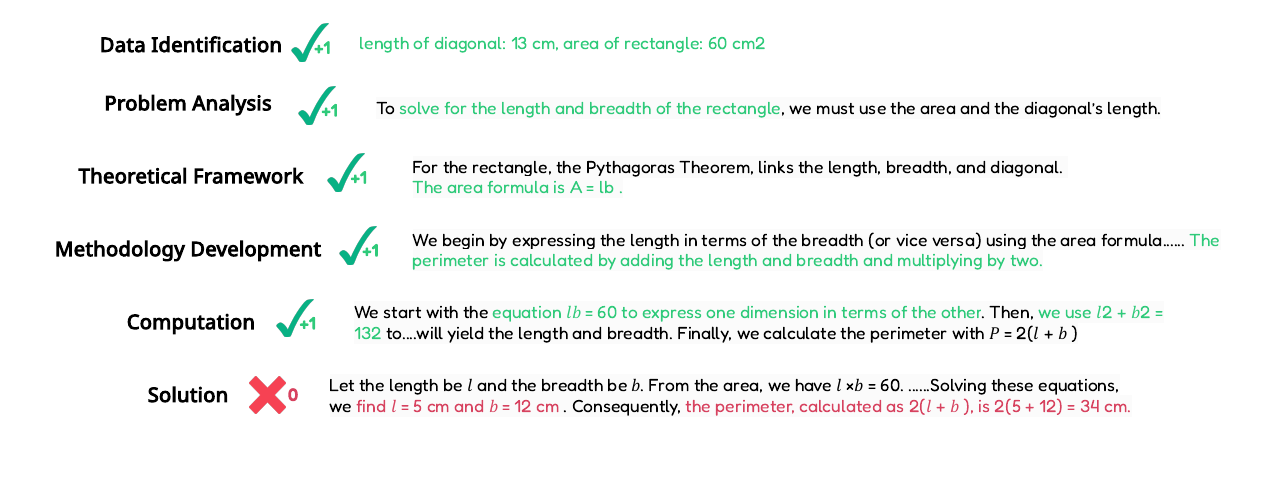}
    \caption{The binary ‘Correct’ or ‘Incorrect’ evaluative approach of existing benchmarks is inadequate to examine the depth and precision of the multi-step reasoning process. To this end, We propose a CoT evaluation strategy to thoroughly assess their mathematical CoT skills in visual contexts, involving Multimodal LLMs}
    \label{fig:coteval}
\end{figure*}

\section{Results and Discussion}
The base model LLaVA 1.6, when finetuned on the PGPS9K dataset, consistently demonstrated low performance with accuracy rates ranging from 7.33\% to 8.66\%. However, significant improvements were noted when the same base model was fine-tuned on our dataset, where the accuracy increased to a range between 22.66\% and 24.66\% across different configurations of the Vicuna and Mistral models. This highlights the effectiveness of the dataset employed for finetuning in enhancing model accuracy.
The GLLaVA models, both in 7B and 13B configurations, show remarkable improvement when fine-tuned on our dataset compared to training on GeoQA++. The GLLaVA 7B model fine-tuned on our dataset achieved an accuracy of 17.33\%, a notable increase from the 10.66\% accuracy obtained with the GeoQA++ dataset. Similarly, the GLLaVA 13B model's accuracy jumped from 22.66\% when trained on GeoQA++ to 25.33\% on our dataset. 
The higher capacity models such as Gemini and GPT-4 show superior performance with accuracies of 35.33\% and 56.66\%, respectively. This indicates that larger models potentially benefit more from increased parameters, allowing for better generalization across complex tasks like geometric problem-solving.

The results indicate that human-annotated captions generally enhance model performance more significantly than captions generated by the Gemini model. For instance, the GLLaVA 7B FT model saw an increase in accuracy from 21.33\% with Gemini captions to 24.66\% with human captions. This suggests that the quality and relevance of human-generated captions are superior. Comparing different model architectures, the Gemini + Human configuration outperformed the GLLaVA-based setups, achieving an accuracy of 29.33\%. Furthermore, results from Table \ref{tab: experiment_4} suggest the effectiveness of RAG in improving the model's ability to tackle complex problems. Despite these advancements, a notable gap remains in the models' ability to utilize theorem-based knowledge effectively to solve problems.

\section{Conclusion and Future Scope:}
In this paper, we introduced the GPSM4K dataset, a multi-modal collection of geometry problems consisting of around 4,000 problems derived from grades 7-12 mathematics textbooks. The dataset, featuring both numerical-answer and theorem-proving questions, is designed to challenge and evaluate the capabilities of modern Large Vision Language Models (LVLMs) in handling complex geometric queries. We will also make the dataset publicly available to encourage further research and collaboration in the field. Furthermore, the results highlight the significant role of dataset choice and caption quality in enhancing model performance on geometry-based visual question answering. While the GPSM4K dataset proves effective, there remains a performance gap that suggests further room for improvement, particularly in how models utilize theorem-based knowledge. Future research should focus on refining datasets to include more theorem-rich content and on developing advanced techniques for generating captions that can encapsulate more detailed geometric information. This direction is likely to bridge the current gap between LLMs and Vision Models and enhance the overall efficacy of models in geometric problem-solving.

\section{Ethics Statement}
The datasets utilized for training and testing the LLM were obtained from publicly available repositories. We acknowledge the possibility of bias present in language model training datasets. The data was sourced from NCERT textbooks, which are extensively used in high schools throughout India, thus reflecting a wide demographic within the country.

\section{Limitations}
One significant limitation of our study is the scale of the dataset. While the GPSM4K dataset introduces 1,438 original geometric problems augmented to a total of 4,440 entries, this size is relatively small compared to larger datasets such as PGPS9K and Geo170K. The limited scale of GPSM4K may constrain the ability to train robust models capable of generalizing across a broader range of geometric problems. This limitation underscores the need for future work to expand the dataset, ensuring a more diverse and extensive collection of geometry problems to enhance the training and evaluation of Large Vision Language Models (LVLMs).
Future studies should involve more controlled and comparable experimental conditions and utilize multiple datasets for fine-tuning to ensure more rigorous and persuasive conclusions.


\bibliographystyle{ACM-Reference-Format}
\bibliography{software}

\clearpage
\appendix
\section{Appendix}

The following appendix provides a detailed overview of the our dataset sample. Some samples from this dataset are shown above,illustrating the key features and data points included. The dataset encompasses two augementated questions and its re generated solutions , which are essential for the analysis conducted in this step by step reasoning. Below, additional samples are presented to give a comprehensive understanding of the dataset's structure and content. \\
    \textbf{Question 2:} \\
In a circle, the diameter is $52 \, \mathrm{cm}$ and the length of a chord is $48 \, \mathrm{cm}$. Determine the distance of the chord from the center. \\

\textbf{Solution 2:} \\
Given diameter $= 52 \, \mathrm{cm}$. \\
So, radius $= \frac{52}{2} = 26 \, \mathrm{cm}$. \\
Hence, $\mathrm{PC} = 26 \, \mathrm{cm}$. \\
Also, length of chord $\mathrm{CD} = 48 \, \mathrm{cm}$. \\

In $\triangle \mathrm{PMC}$, $\angle \mathrm{PMC} = 90^{\circ}$. \\

Using Pythagoras theorem:
\[
\mathrm{PC}^2 = \mathrm{PM}^2 + \mathrm{CM}^2
\]

Thus,
\[
26^2 = \mathrm{PM}^2 + \left(\frac{1}{2} \times 48\right)^2
\]
\[
676 = \mathrm{PM}^2 + 576
\]
\[
\mathrm{PM}^2 = 100
\]
\[
\mathrm{PM} = 10 \, \mathrm{cm}
\]

So, the distance of the chord from the center is $10 \, \mathrm{cm}$.

    \bigskip
    \hrule
    \bigskip

    \textbf{Solution Regeneration}
    \medskip
    \hrule
    \medskip
    \textbf{Question (Original):} Diameter of a circle is $26 \mathrm{~cm}$ and the length of the chord of a circle is $24 \mathrm{~cm}$. Find the distance of the chord from the center.
    \textbf{Solution:}
    \begin{itemize}
    \item \textbf{Data Identification}:
        \begin{itemize}
            \item Diameter of the circle: $26 \mathrm{~cm}$
            \item Length of the chord: $24 \mathrm{~cm}$
        \end{itemize}
        
    \item \textbf{Problem Analysis}:
        \begin{itemize}
            \item We need to find the distance from the center of the circle to the chord.
        \end{itemize}
        
    \item \textbf{Theoretical Framework}:
        \begin{enumerate}
            \item In a circle, the radius is half the diameter.
            \item Perpendicular drawn from the center of the circle to the chord bisects the chord.
            \item Pythagoras theorem: In a right triangle, the square of the length of the hypotenuse (the side opposite the right angle) is equal to the sum of the squares of the lengths of the other two sides.
        \end{enumerate}
        
    \item \textbf{Methodology Development}:
        \begin{itemize}
            \item Let:
                \begin{itemize}
                    \item $PC$ be the radius of the circle.
                    \item $PM$ be the distance from the center to the chord.
                    \item $CM$ be the distance from the chord to the midpoint of the chord.
                \end{itemize}
            \item From the given data and the properties of circles:
                \begin{enumerate}
                    \item Radius $PC = \frac{1}{2}$ Diameter = $\frac{26}{2} = 13 \mathrm{~cm}$.
                    \item $CM = \frac{1}{2} \times CD$ (Given that the perpendicular drawn from the center of the circle to the chord bisects the chord).
                    \item By Pythagoras theorem: $PC^2 = PM^2 + CM^2$.
                \end{enumerate}
        \end{itemize}
        
    \item \textbf{Computation}:
        \begin{enumerate}
            \item Calculate the radius $PC = 13 \mathrm{~cm}$.
            \item Calculate $CM = \frac{1}{2} \times CD = \frac{1}{2} \times 24 = 12 \mathrm{~cm}$.
            \item Use Pythagoras theorem: $PC^2 = PM^2 + CM^2$.
                \begin{itemize}
                    \item $13^2 = PM^2 + 12^2$
                    \item $169 = PM^2 + 144$
                    \item $PM^2 = 169 - 144 = 25$
                    \item $PM = \sqrt{25} = 5 \mathrm{~cm}$.
                \end{itemize}
        \end{enumerate}
        
    \item \textbf{Solution}:
        \begin{itemize}
            \item The distance from the center of the circle to the chord is $5 \mathrm{~cm}$.
        \end{itemize}
\end{itemize}

\textbf{Question (Augmented 1):} \\
The diameter of a circle is $13 \, \mathrm{cm}$, and the length of a chord of the circle is $12 \, \mathrm{cm}$. Find the distance of the chord from the center. \\
\textbf{Solution:}
\begin{itemize}
    \item \textbf{Data Identification:}
    \begin{itemize}
        \item Diameter of the circle: $13 \, \mathrm{cm}$.
        \item Length of chord: $12 \, \mathrm{cm}$.
    \end{itemize}
    
    \item \textbf{Problem Analysis:}
    \begin{itemize}
        \item We need to find the distance of the chord from the center of the circle.
    \end{itemize}
    
    \item \textbf{Theoretical Framework:}
    \begin{itemize}
        \item \textbf{Pythagoras' Theorem:} In a right triangle, the square of the length of the hypotenuse (the side opposite the right angle) is equal to the sum of the squares of the lengths of the other two sides.
    \end{itemize}
    
    \item \textbf{Methodology Development:}
    \begin{itemize}
        \item Let's denote:
        \begin{itemize}
            \item Radius of the circle as $r$.
            \item Distance from the center to the chord as $PM$.
            \item Length of the chord as $CD$.
            \item Half the length of the chord as $CM$.
        \end{itemize}
        \item Given:
        \begin{itemize}
            \item Diameter of the circle = $13 \, \mathrm{cm}$, so $r = \frac{13}{2} = 6.5 \, \mathrm{cm}$.
            \item $CD = 12 \, \mathrm{cm}$.
        \end{itemize}
        \item We can find $CM$ using $CD$, as $CM = \frac{1}{2} CD = 6 \, \mathrm{cm}$.
        \item In $\triangle PMC$, we have:
        \begin{itemize}
            \item $\angle PMC = 90^\circ$ (as $PM$ is perpendicular to $CD$).
            \item $PC = r = 6.5 \, \mathrm{cm}$.
            \item $CM = 6 \, \mathrm{cm}$.
        \end{itemize}
        \item Using Pythagoras' theorem: 
        $$PC^2 = PM^2 + CM^2$$
    \end{itemize}
    
    \item \textbf{Computation:}
    \begin{itemize}
        \item Substitute the given values:
        $$6.5^2 = PM^2 + 6^2$$
        $$42.25 = PM^2 + 36$$
        $$PM^2 = 42.25 - 36 = 6.25$$
        $$PM = \sqrt{6.25} = 2.5 \, \mathrm{cm}$$
    \end{itemize}
    
    \item \textbf{Solution:}
    \begin{itemize}
        \item The distance of the chord from the center is $2.5 \, \mathrm{cm}$.
    \end{itemize}
\end{itemize}

\textbf{Question (Augmented 2):} \\
In a circle, the diameter is $52 \, \mathrm{cm}$ and the length of a chord is $48 \, \mathrm{cm}$. Determine the distance of the chord from the center. \\
\textbf{Solution:}
\begin{itemize}
    \item \textbf{Data Identification:}
    \begin{itemize}
        \item Diameter of the circle: $52 \, \mathrm{cm}$
        \item Length of chord: $48 \, \mathrm{cm}$
    \end{itemize}
    
    \item \textbf{Problem Analysis:} \\
    We need to find the distance of the chord from the center of the circle.
    
    \item \textbf{Theoretical Framework:}
    \begin{enumerate}
        \item In a circle, the radius is perpendicular to the chord at the midpoint of the chord.
        \item Pythagoras theorem states that in a right triangle, the square of the length of the hypotenuse (the side opposite the right angle) is equal to the sum of the squares of the lengths of the other two sides.
    \end{enumerate}
    
    \item \textbf{Methodology Development:}
    \begin{enumerate}
        \item Let $\mathrm{PM}$ be the distance from the center $P$ to the midpoint $M$ of the chord $\mathrm{CD}$.
        \item Using Pythagoras theorem in triangle $\triangle \mathrm{PMC}$:
        \[
        \mathrm{PC}^2 = \mathrm{PM}^2 + \mathrm{CM}^2
        \]
        where $\mathrm{PC}$ is the radius of the circle, and $\mathrm{CM}$ is half the length of the chord.
        \item We're given the radius and the length of the chord, so we can solve for $\mathrm{PM}$.
    \end{enumerate}
    
    \item \textbf{Computation:}
    \begin{itemize}
        \item Given:
        \begin{itemize}
            \item Diameter $= 52 \, \mathrm{cm}$ (So, radius $= \frac{52}{2} = 26 \, \mathrm{cm}$)
            \item Length of chord $= 48 \, \mathrm{cm}$
        \end{itemize}
        \item Using the formula:
        \[
        \mathrm{PC}^2 = \mathrm{PM}^2 + \mathrm{CM}^2
        \]
        \item Substituting the values:
        \[
        26^2 = \mathrm{PM}^2 + \left(\frac{1}{2} \times 48\right)^2
        \]
        \[
        676 = \mathrm{PM}^2 + 576
        \]
        \[
        \mathrm{PM}^2 = 100
        \]
        \[
        \mathrm{PM} = 10 \, \mathrm{cm}
        \]
    \end{itemize}
    
    \item \textbf{Solution:} \\
    The distance of the chord from the center is $10 \, \mathrm{cm}$.
\end{itemize}

\end{document}